\documentclass[journal]{IEEEtran}
\usepackage{times}
\usepackage{epsfig}
\usepackage{graphicx}
\usepackage{amsmath}
\usepackage{amssymb}
\usepackage{siunitx}
\usepackage{xcolor}
\usepackage{amssymb}
\usepackage{algorithm}
\usepackage[noend]{algpseudocode}
\usepackage{multirow}
\usepackage{booktabs}

\pdfoutput=1
\DeclareMathOperator*{\argmax}{arg\,max}

\begin{document}

\title{Detection and Localization of Facial Expression Manipulations}

\author{Ghazal Mazaheri,~\IEEEmembership{Member,~IEEE,}
        Amit K. Roy-Chowdhury,~\IEEEmembership{Fellow,~IEEE,}
\thanks{Ghazal Mazaheri is with the Department
of Computer Science and Engineering, University of California, Riverside, CA, USA. Amit K. Roy-Chowdhury are with the Department
of Electrical and Computer Engineering, University of California, Riverside, CA, USA. E-mails: (gmaza002@ucr.edu, amitrc@ece.ucr.edu)}
}

\markboth{IEEE TRANSACTIONS ON INFORMATION FORENSICS AND SECURITY,~Vol.X, No.X, 2021}
{Mazaheri\MakeLowercase{\textit{et al.}}: Detection and Localization of Facial Expression Manipulations}

\maketitle
\begin{abstract}
Concern regarding the wide-spread use of fraudulent images/videos in social media necessitates precise detection of such fraud. The importance of facial expressions in communication is widely known, and adversarial attacks often focus on manipulating the expression related features. 
Thus, it is important to develop methods that can detect manipulations in facial expressions, and localize the manipulated regions.
To address this problem, we propose a framework that is able to detect manipulations in facial expression using a close combination of facial expression recognition and image manipulation methods. With the addition of feature maps extracted from the facial expression recognition framework, our manipulation detector is able to localize the manipulated region. We show that, on the Face2Face dataset, where there is abundant expression manipulation, our method achieves over 3\% higher accuracy for both classification and localization of manipulations compared to state-of-the-art methods. In addition, results on the NeuralTextures dataset where the facial expressions corresponding to the mouth regions have been modified, show 2\% higher accuracy in both classification and localization of manipulation. We demonstrate that the method performs at-par with the state-of-the-art methods in cases where the expression is not manipulated, but rather the identity is changed, thus ensuring generalizability of the approach. 
\end{abstract}


\IEEEpeerreviewmaketitle

\section{Introduction}
 Facial expressions are critical in communicating our thoughts, ideas and emotions, and in responding to each other emotionally and physically. Such responses may even be elicited without the receiver's conscious awareness \cite{Frith_Chris}. With command over facial expressions, a person may convince others to believe in ideas they wish to convey. Due to the power of facial expressions in person-to-person communication, it is critical to determine if the facial expressions in an image/video are the individual's original expressions or manipulated by an external agent.

\begin{figure}[t]
\includegraphics[width=\linewidth]{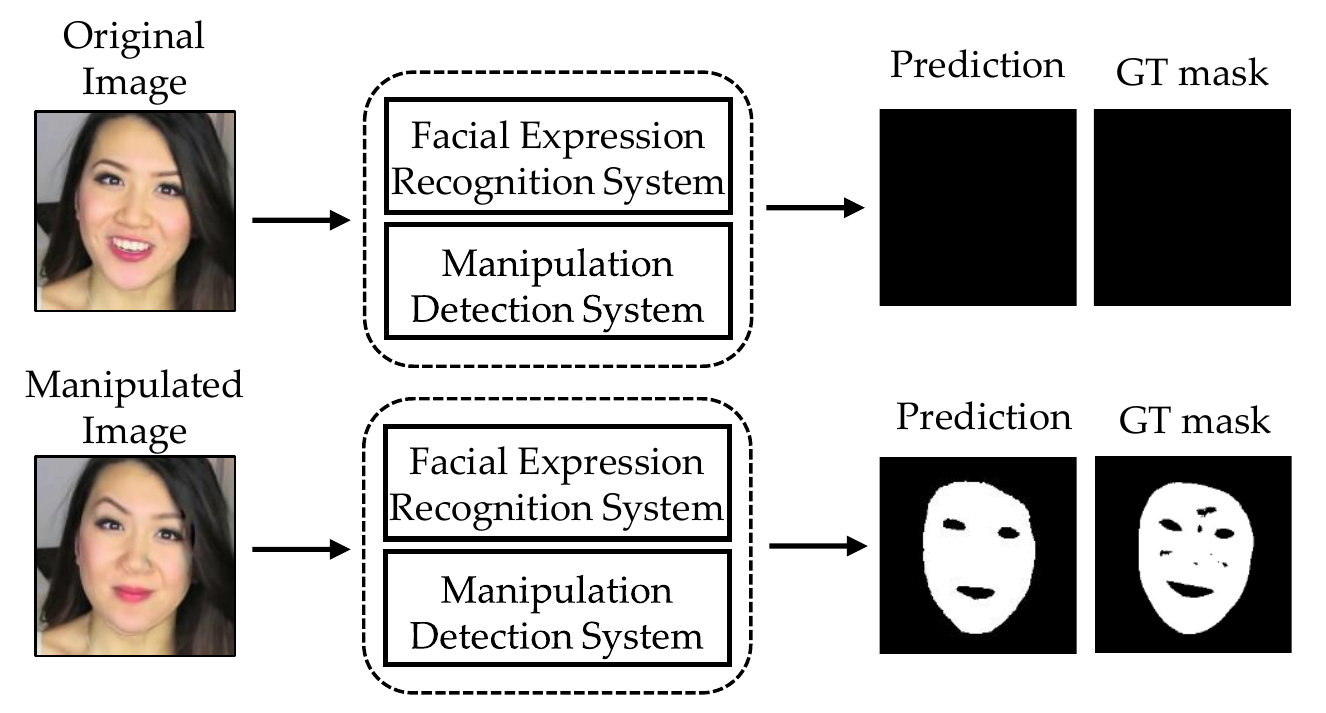}
\caption{Our overall framework for detection of facial expression manipulation and localization. Original image, manipulated version and their ground-truth (GT) masks are from FaceForensics++ dataset \cite{Face2face}}.
\label{fig1}
\end{figure}

\begin{figure*}[t]
\includegraphics[width=\textwidth]{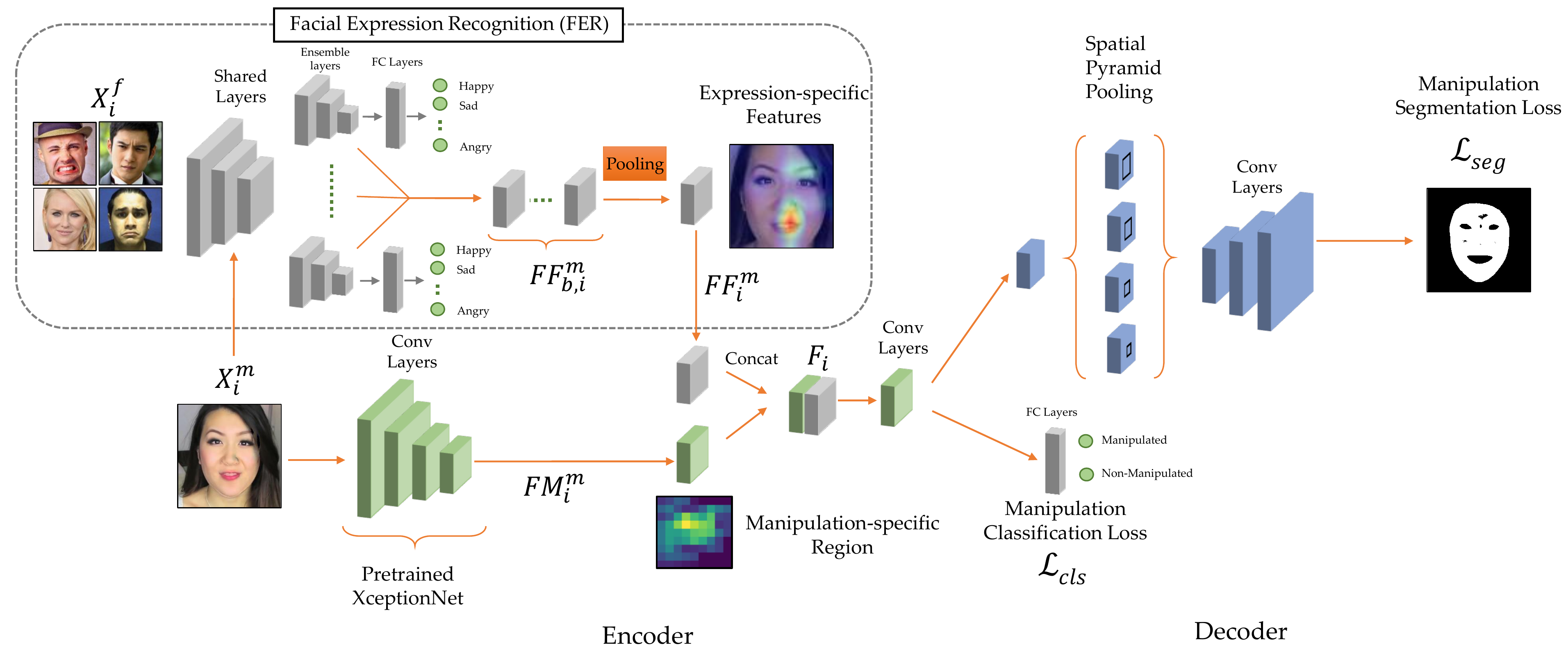}
\caption{This figure represents our proposed approach for facial expression manipulation detection and localization. Extracted features from FER System $(FF_i^m)$, along with the ones from manipulation detection stream $(FM_i^m)$, are fed into the decoder for pixel-wise localization of the manipulated region. Notation is described in the text. The details are explained is Sec. \ref{FER} and \ref{EnDe}} \label{fig2}
\end{figure*}

In this work, \emph{we focus on the problem of detecting facial expression manipulation}, such as those produced by current state-of-the art face manipulation algorithms for facial expression change, e.g. Face2Face \cite{Face2face} and NeuralTextures \cite{NT}. We call this as the Expression Manipulation Detection (EMD) system. Facial expression changes in the Face2Face dataset is a result of a facial reenactment system that transfers the expressions of a source video to a target video while maintaining the identity of the target person. NeuralTextures \cite{NT} reenacts face motions of an input video to a target video mainly affecting the regions around the mouth. We hypothesize that to detect facial expression manipulations, recognition of the expression would be helpful. Fig. \ref{fig1} presents our idea of using facial expression recognition to guide the manipulation detection procedure. As can be seen from the figure, the main manipulations appear in the parts of the face which are important for expression change, such as, regions around eyebrows and mouth which are critical regions for facial expressions.

In order to utilize prominent features corresponding to facial expression, we adapt existing state-of-the-art Facial Expression Recognition (FER) systems to our face manipulation detection framework (see Sec. \ref{FER} for details). In this work, we utilize Ensemble with Shared Representations (ESR) \cite{siqueira2020efficient} as the FER system. Feature maps from the penultimate layer of FER systems contain important information regarding facial expressions in faces~\cite{Zhou_2016}, which we exploit to improve over state-of-art manipulation detection methods. 

We also want our method to be applicable when faces are manipulated in other ways, such as, by changing identity. In these situations, we want to ensure that our method does not reduce existing performance. Thus, we  also  demonstrate that the method performs at-par with the state-of-the-art methods in cases where the expression is not manipulated, but rather the identity is changed, thus ensuring generalizability of the approach.

\subsection{Framework Overview}
A pictorial flow of our facial expression manipulation detection (EMD) framework is presented in Fig. \ref{fig2}.
The proposed method utilizes a two-stream network for manipulation detection. One stream (FER) is responsible for extracting important information for facial expressions. The feature maps from the last layer of FER stream provide information about the facial regions that encode the expression information. The second stream is an encoder-decoder architecture which is responsible for manipulation detection. The encoder projects the image to a lower dimensional space, where the features from the FER system are combined and then a decoder is used to predict the manipulated regions (if any) of the facial image. 

Our face expression recognition system uses ESR \cite{siqueira2020efficient} to extract expression relevant features. An ESR consists of two building blocks. 1. The base of the network is an array of convolutional layers for low- and middle-level feature learning. (Shared Layers in Fig. \ref{fig2}). 2. These informative features are then shared with independent convolutional branches (Ensemble Layers) that constitute the ensemble. We choose the feature maps from the Ensemble branch which has the most frequent expression prediction category with the highest probability score (see Sec. \ref{FER}). The penultimate layer feature map of the FER system contains features which are discriminative to detect relevant portions of the image specific for expression. The second stream, i.e., the encoder-decoder architecture is an Encoder-Decoder with Atrous Separable Convolution. It has a similar architecture in the encoder part as the XceptionNet \cite{Rossler_2019_ICCV} architecture with minor changes (see Fig. \ref{fig2}). The decoder then takes the latent space features from the encoder and FER system combined and projects it using DeepLabv3+ \cite{deeplab} (which employs the spatial pyramid pooling module with the encoder-decoder structure) for manipulation localization. (see Fig. \ref{fig2}). 
\vspace{-0.3cm}
\paragraph{Main contributions.}
We propose a novel approach for facial expression manipulation detection leveraging upon a facial expression recognition system.
This leads to higher performance in forgery detection where the facial expression is manipulated, as well as localizing the regions that have been manipulated. This is achieved without drop in performance in the case where the identity is manipulated, thus ensuring generalizability across different kinds of manipulation. Our method leads to more than 3\% improvement in both manipulation classification and localization over the state of the art on the Face2Face dataset \cite{Face2face} where the expressions are manipulated. We also show the effectiveness of our method by presenting the results on NeuralTextures dataset \cite{NT} where the facial expressions corresponding to the mouth regions have been modified. On NeuralTextures dataset, we achieve 2\% higher accuracy for classification and localization of manipulation.  Our method achieves almost the same result as in the state-of-the-art methods in DeepFake dataset \cite{Deepfake} where identity, rather than expression, is manipulated.

 

\section{Related Works}
Multimedia forensics aims to ensure authenticity, origin, and provenance of an image or video. In recent years, there has been a variety of works in 1. forgery classification where a proposed method recognizes whether or not an image or video is manipulated, and 2. forgery localization which highlights the exact position of manipulated region \cite{tutorial}. We will briefly survey existing work in both the mentioned categories, as well as facial expression recognition. There is no work that specifically focuses on the problem of detection and localization of facial expression manipulations.

\subsection{Forgery Classification}
In forgery classification area, there has been a variety of works in image manipulation detection \cite{bayar,denoising,Croping,rahmouni,jawad, mantra, rich} or fake faces classification in videos \cite{Rossler_2019_ICCV, Afchar2018MesoNetAC,Dang_2020_CVPR,Li_2020_CVPR}.
The very first works aim to detect wide range of manipulations in images including object removal, copy-move, and splicing using handcrafted features that capture expected statistical or physics-based features which occur during image formation. Authors in \cite{Croping} propose an approach for detecting copy-paste and composite forgery in JPEG images. Work in \cite{denoising} investigates the inﬂuence of denoising on PRNU-based forgery detection. Inspired by the success of deep neural networks in different visual recognition tasks, deep learning-based approaches have been popular choices for image forgery detection. Some recent deep learning-based methods such as convolutional neural networks (CNN) have been applied to detect/classify image manipulations \cite{bayar, jawad, mantra, rich}.

Manipulation of faces in images/videos has been in the news lately.
Manipulation detection in faces is challenging since exiting manipulation techniques leave almost no visual traces. In \cite{Dang_2020_CVPR} authors propose to utilize an attention mechanism to process and improve the feature maps for the classification task. Four main categories of face manipulation techniques include Face2Face \cite{Face2face}, DeepFake \cite{Deepfake}, FaceSwap \cite{faceswap} and  NeuralTextures \cite{neuraltext}. The most comprehensive dataset containing all four manipulation techniques (FaceForensics++) has been introduced in \cite{Rossler_2019_ICCV}. Generative networks play an important role in producing manipulated images and videos. Work in \cite{spaGen} proposes a new model to perform multiple facial attributes manipulation with one-input multi-output architecture.

To detect face manipulation in videos, some approaches utilize video temporal features as tampering of individual frames in videos causes inconsistency. The work in \cite{RNN_CNN} uses CNN as feature extractor and LSTM to capture video temporal features. Some other works use physiological signals, like eye blinking in \cite{Eye} and head movements in \cite{HeadPose}, that are not well presented in the synthesized fake videos. Instead of using temporal features, authors in \cite{Rossler_2019_ICCV, Afchar2018MesoNetAC, Li_2019_CVPR_Workshops, Nguyen2019MultitaskLF} proposed methods which utilize extracted images from different frames of a video. Work in \cite{speak} proposes a visual speaker authentication scheme based on the deep convolutional neural network (DCNN) to defend against DeepFake attacks. 

\subsection{Forgery Localization}
Localizing the exact position of manipulated regions in an image or video provides critical additional information. There has been a variety of works that attempted to segment out tampered regions \cite{Deepmatch, Bappy_2017_ICCV, Zhou_2018_CVPR}. Early works \cite{Ryu2013RotationIL, Ferrara2012ImageFL, Bianchi2012ImageFL} reveal the tampered regions using traditional image processing-based approaches. Researchers in \cite{Bappy_2017_ICCV, Liu_2018_ECCV_Workshops,Bunk:2017, Zhang2018BoundarybasedIF} exploit machine learning techniques in order to classify if a patch is manipulated or not. 

 Authors in~\cite{Zhou_2018_CVPR} use object detection method proposed in~\cite{R-CNN} to identify fake objects. Unlike \cite{Zhou_2018_CVPR} which utilizes bounding box to coarsely localize manipulated object, \cite{jawad} adopt a segmentation approach to segment out manipulated regions by classifying each pixel (manipulated/non-manipulated). Semantic segmentation approaches are suitable for fine-grained localization of tampered regions in an image. A typical semantic segmentation approach focuses on segmenting all meaningful regions (objects). However, a segmentation approach for localization of image manipulation needs to focus only on the possible tampered regions which bring additional challenges to an existing challenging problem. To localize tampered regions, \cite{jawad} used an LSTM Encode-Decoder architecture.

Fake face segmentation is one of the recent challenges which has not yet been addressed extensively. Some of the proposed methods may have high performance in face manipulation detection \cite{Rossler_2019_ICCV} but do not address the task of segmenting the manipulated region. Multi-tasking approaches are promising in the combined classification and segmentation task. Work in \cite{Nguyen2019MultitaskLF} uses Y-shape architecture to classify manipulated videos and segment tampered faces simultaneously. In our proposed method, in addition to manipulation classification and segmentation stream, we add another stream as face expression recognition which operates jointly with the manipulation stream in order to exploit necessary information in faces. This improves the performances in both classification and segmentation.

\subsection{Facial Expression Recognition}
The development of machine learning and the advent of deep learning have significantly improved the research of FER. There have been variety of works in literature which obtain high performance for facial expression recognition framework \cite{Hassani2017SpatioTemporalFE, He2015DeepRL, Li_2017_CVPR,siqueira2020efficient, Yang_2018_CVPR, Zeng_2018_ECCV}.  The careful design of local to global feature learning with a convolution, pooling, and layered architecture produces a rich visual representation, making CNN a powerful tool for facial expression recognition. Research challenges such as the Emotion Recognition in the Wild (EmotiW) and Kaggle’s Facial Expression Recognition Challenge suggest the growing interest of the community in the use of deep learning for the solution of this problem. 

To have more accurate FER, the networks become deeper and deeper in order to deal with more complex classifications. Also, attention mechanisms are introduced in many networks to improve facial expression recognition. Authors in \cite{SUN201812} proposed a facial expression recognition network with the visual attention mechanism. Work in \cite{siqueira2020efficient} is one of the most recent in this area showing promising results on facial expression datasets using shared and ensemble layers.

\section{Methodology}
In this section, we present our framework for facial expression manipulation detection and localization. We start with a formal description of the problem statement followed by the two streams of our framework - Facial Expression Recognition (FER) stream and the encoder-decoder based manipulation detection stream which receives information from FER for better detection. 

\subsection{Problem Statement}
Consider we have a dataset of tuples $\mathcal{X}^M=\{(X_i^m, M_i^m, y_i^m)\}_{i=1}^N$, where $X_i^m \in \mathbb{R}^{H \times W \times 3}$ is a 2D image of faces, $M_i \in \mathbb{R}^{H \times W}$ is 2D binary mask of manipulated regions, and $y
_i \in \{0,1\}$ is an indicator whether $y_i^m$ is manipulated or not. Given such a dataset, our main goal is to learn a model that would be able to classify a test image to be either manipulated or not, and more importantly, localize portions of the image which are manipulated.

To identify manipulations in facial expressions, we need to focus on regions specific for expressions; thus, we utilize an auxiliary task of Facial Expression Recognition (FER). We use a dataset of tuples $\mathcal{X}_F=\{(X_i^f, y_i^f)\}_{i=1}^{N'}$, where $X_i^f \in \mathbb{R}^{H \times W \times 3}$ and $y_i^f \in \{1, \dots, C\}$, where $C$ is the number of facial expression category. Note that we use the superscripts $m$ and $f$ to denote data points from the manipulation set $\mathcal{X}^M$ and facial expression set $\mathcal{X}^F$ respectively. 

\vspace{0.3cm}

\subsection{Algorithm Overview}
Our EMD system consists of two main parts including 1. FER, 2. encoder-decoder.
We train the FER module using the dataset $\mathcal{X}^F$. To train the encoder-decoder architecture for manipulation detection, our framework takes information from the FER module. However, we pass the images in $\mathcal{X}^M$ through both the streams - 1. an encoder to obtain features necessary to detect manipulations, 2. FER module to obtain features specific to facial expressions. We combine these features in the latent space and then pass them through a decoder to spatially localize the manipulated regions.

\vspace{0.3cm}

\subsection{Facial Expression Recognition}
\label{FER}
We utilize one of the state-of-the-art methods for facial expression recognition proposed in \cite{siqueira2020efficient} as a pre-trained model for recognizing the facial expressions in our framework. 

FER system presented in \cite{siqueira2020efficient} consists of two building blocks. 1. The base of the network (shared layers in Fig. \ref{fig2}) is an array of convolutional layers for low- and middle-level feature learning, 2. These informative features are then shared with independent convolutional branches that constitute the ensemble (Ensemble layers in Fig. \ref{fig2}). From this point, each branch can learn distinctive features while competing for a common resource - the shared layers. This competitive training emerges from the minimization of a combined loss function defined as the summation of the loss functions (cross-entropy loss) of each branch as follows: 

\begin{equation}
\mathcal{L}_{FER}=\frac{\num{1}}{N'}\sum\limits_b\sum\limits_i{\sum\limits_c -{y_{i,c}^f\log(p_{b,i,c}^f)}}
\end{equation}
where given an image $X_i^f$, $p_{b,i,c}^f$ is a probability mass function over the $C$ facial expression categories for branch $b$. $N'$ is total number of images in the dataset.

After training the FER system, we use the pre-trained models to predict the expression category of manipulated images and extract the feature maps needed to be combined with manipulation detection stream. The reason is that, for manipulation detection task, feature maps which highlight the discriminative image regions important for expression recognition are useful for manipulation detection of images where facial expressions are tampered such as manipulation in Face2Face dataset. Therefore, our FER architecture provides useful expression and location-aware features needed for the manipulation detection task.

In the Facial Expression Recognition system, we pass an image $X_i^m$ through the shared convolutional network $Conv$ and an ensemble of $B$ convolutional networks $\{\mathcal{E}_b\}_{b=1}^{B}$ to generate $B$ different feature maps of the facial expression. For the $b^{th}$ ensemble convolutional branch, the feature map corresponding to $X_i^m$ is,
\begin{equation}
    FF_{b,i}^m= \mathcal{E}_b(Conv(X_i^m)).
\end{equation} 
We use a classifier $\mathcal{M}$ to infer class probabilities for the $C$ expression classes. For $b^{th}$ branch network and $i^{th}$ image $X_i^m$, we infer the class probability vector, $\boldsymbol{\rho}_{b,i} = [\rho_{b,i,1}, \dots, \rho_{b,i,C}]$. Here, $\rho_{b,i,j} = \mathcal{M}(FF_{b,i}^m,c_j)$ indicates the detection probability of class $c_j$ for branch $b$ with input image $X_i^m$ . Therefore, the detected expression class of an image from $b^{th}$ convolutional branch is,
\begin{equation}
    c_{det}^b = \argmax_{c_j \in \{c_1, \dots, c_C\}} \mathcal{M}(FF_{b,i}^m,c_j).
\end{equation}
Considering the detection of all the branches, most frequent detected class for an image is, 
\begin{equation}
    c_{freq} = mode(\{c_{det}^1, \dots, c_{det}^B \})     
\end{equation}
The feature map from the branch in FER network that results in highest detection probability for the frequently detected class $c_{freq}$ is pooled for manipulation detection task. So, the pooled feature map is,
\begin{equation}
    FF_i^m = \argmax_{FF_{b,i}^m \in \{FF_{1,i}^m, \dots, FF_{B,i}^m\} 
    } \mathcal{M}(FF_{b,i}^m,c_{freq})
\end{equation}

Note, as will be discussed subsequently, we use the feature map $FF_i^m$ after the convolutional layers as auxilliary input to the encoder-decoder stream for manipulation detection. As illustrated later in Fig. \ref{fig7}, we also obtain the class activation maps (CAMs), for visualization purpose following \cite{Zhou_2016}.

\subsection{Encoder-Decoder for Manipulation Detection and Segmentation}
\label{EnDe}
Encoder-decoder networks using CNN architecture have been extensively used in deep learning literature, specifically for  semantic object segmentation. Following the literature, we adopt Encoder-Decoder with Atrous Separable Convolution architecture \cite{deeplab} known as deeplabv3+ for manipulation detection and segmentation as the task of localizing manipulation regions is similar to semantic segmentation task. The encoder in this architecture is an XceptionNet which has the best accuracy in manipulation detection among the state-of-arts. The Spatial Pyramid Pooling (SPP) module consists of depthwise separable convolution layers resulting in a faster and stronger encoder-decoder network.

Given an image $X_i^m$, we pass it through the encoder to obtain $FM_i^m$ from one before the last convolutional layers. As we are interested in detecting manipulations in expression, we inject features from the facial expression recognition stream into the encoder-decoder manipulation detection stream. To do that, we also pass $X_i^m$ through FER and obtain features $FF_i^m$. We then concatenate both the feature maps as $F_i=FM_i^m$$\oplus$$FF_i^m$ and pass the concatenated feature maps through remaining layers of encoder to obtain latent space features. As shown in Fig. \ref{fig2}, we have two loss functions for classification $(\mathcal{L}_{cls})$ and segmentation $(\mathcal{L}_{seg})$. We use cross-antropy loss function for classification task defined as follows:
\begin{equation}
\mathcal{L}_{cls}=\frac{1}{N}\sum_i {\left\|y_{i}^m\log(a_i)+(1-y_{i}^m)\log(1-a_i)\right\|_1},
\end{equation}
where $a_i$ is the output of binary classification which determines whether or not an image is manipulated. 

Next, the decoder takes  the latent space features as the input.  Consider that $S_i \in \mathbb{R}^{H \times W}$ is the spatial manipulation segmentation output. We compute the segmentation loss function to measure the agreement between the segmentation mask and the ground-truth mask as follows:
\begin{equation}
\mathcal{L}_{seg}=\frac{1}{N}\sum_i {\left\|M_{i}^m\log(S_i)+(1-M_{i}^m)\log(1-S_i)\right\|_1},
\end{equation}
Note that each pixel in $S_i$ lies in between $0$ and $1$ depicting the probability of it being manipulated or not. 

The total loss function we optimize to learn the encoder decoder architecture is as follows:
\begin{equation}
    \mathcal{L}_{MANI} = \mathcal{L}_{cls} + \mathcal{L}_{seg} 
\end{equation}

\subsection{Overall Algorithm}
Here we discuss the overall training strategy for EMD algorithm. This is presented in Algorithm \ref{alg:train}. Consider that the FER network is parameterized by $\phi$ and the the encoder-decoder for manipulation detection is parameterized by $\theta$. We learn them separately. First we sample images from the facial expression dataset $\mathcal{X}^F$, compute the loss $\mathcal{L}_{FER}$ and update $\phi$ using it. We then sample images from the manipulated images dataset $\mathcal{X}^M$, pass them through both pretrained FER stream and encoder-decoder stream, compute the loss $\mathcal{L}_{MANI}$ and then update $\theta$.

\begin{algorithm}
\caption{Overall EMD Algorithm}
\begin{algorithmic}[1]
 \State \textbf{Inputs:}\hspace{1mm} 1. Expression Recognition Dataset: $\mathcal{X}^F$ \\
 \hspace{12mm} 2. Expression Manipulation Dataset: $\mathcal{X}^M$
 \State \textbf{Output:} Manipulation Detection Network: $\theta$
 \State \textbf{Random Init.:}\\ 
 \hspace{12mm} 1. Facial Expression Recognition Net: $\phi$ \\
 \hspace{12mm} 2. Manipulation Detection Net: $\theta$
 \While{$not converged$}
    \State Mini-batch $B^f=\{X_i^f, y_i^f\}_{i=1}^B\sim\mathcal{X}^F$
    \State Compute: $\mathcal{L}_{FER}(B^f; \phi)$
    \State Update: $\phi \leftarrow \phi - \eta \triangledown_\phi \mathcal{L}_{FER}$
 \EndWhile
 
  \While{$not converged$}
    \State Mini-batch $B^m=\{X_i^m, M_i^m, y_i^m\}_{i=1}^B\sim\mathcal{X}^M$
    \State Compute: $\mathcal{L}_{MANI}(B^m; \theta, \phi)$
    \State Update: $\theta \leftarrow \theta - \eta' \triangledown_\theta \mathcal{L}_{MANI}$
 \EndWhile
\end{algorithmic}
\label{alg:train}
\end{algorithm}

\begin{figure*}
\includegraphics[width=\linewidth]{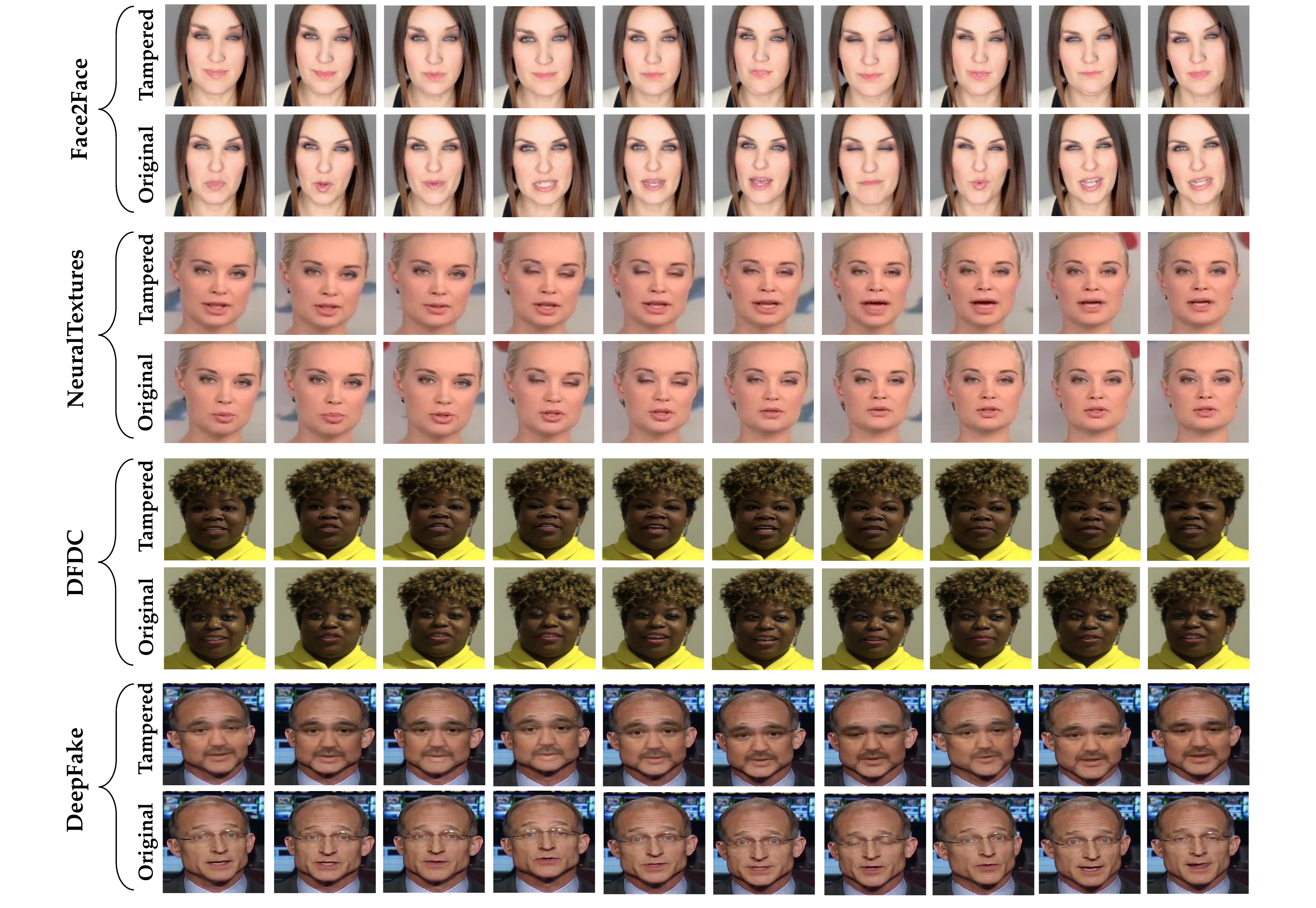}
\caption{Four examples of pristine videos and their manipulated versions from F2F, NT, DFDC and DF datasets. As we can see, facial expression is manipulated in F2F and NT videos while in DFDC and DF datasets identities are swapped.}
\label{fig4}
\end{figure*}

\section{Experiments}
In this section, we perform extensive experiments on three benchmark datasets from FaceForensics++ (FF++) \cite{Faceforensics} to investigate the efficacy of the proposed method. We show results on two datasets where the images correspond to facial expression manipulation and a third dataset where the images undergo an identity change.

\begin{table*}
\centering
\caption{Benchmark Datasets for DeepFake Video Detection. Our approach is applicable to datasets that include facial expression manipulation. Only two datasets (Face2Face and NeuralTextures from Faceforensics++) satisfy that criteria.}
\begin{tabular}{c|c|p{.6cm}p{.6cm}p{.9cm}|c|c|p{1.5cm}p{1.5cm}}
\hline
\multirow{2}{*}{Dataset} & \multirow{2}{*}{Released} & \multicolumn{3}{c|}{\underline{\# Videos}} & \multirow{2}{*}{Real Video Source}& 
\multirow{2}{*}{Method}& 
\multicolumn{2}{c}{\underline{Fake Type}}\\
& & Real & Fake  & Total &  &  & Id Swap & Exp Swap  \\
\hline
\hline
&&&&&&\\
UADFV \cite{HeadPose} & Nov 2018 & \centering 49 & \centering 49 & \centering 98 & YouTube & 1 &\centering \checkmark &\\

DF-TIMIT\cite{timit} & Dec 2018 & \centering 0 & \centering 620 &  \centering 620 & VidTIMIT & 2 & \centering \checkmark & \\

DFD \cite{dfc} & Sep 2019 & \centering 361 & \centering 3070 & \centering 3431 & YouTube & 5 & \centering \checkmark&\\

CelebDF \cite{Celeb} & Nov 2019 & \centering 408 & \centering 795 & \centering 1203 & YouTube & 1 & \centering \checkmark&\\

DFDC \cite{DFDC} & Oct 2019 & \centering 19154 & \centering 99992 & \centering 119146 & Actors & 8 & \centering \checkmark& \\

Deeper Forensics 1.0 \cite{deeperforensics10} & Jan 2020 & \centering 50000 & \centering 10000 & \centering 60000 & Actors& 1 & \centering \checkmark& \\

FaceForensics++ \cite{Faceforensics} & Jan 2019& \centering 1000 & \centering 4000 & \centering 5000 & YouTube & 4 & \centering \checkmark & \centering \checkmark\\
\end{tabular}
\label{table1}
\end{table*}

\subsection{Datasets}
\textbf{FaceForensics++ Dataset.} For our experiments, we used the videos offered by FaceForensics++ \cite{Rossler_2019_ICCV} \footnote{https://github.com/ondyari/FaceForensics}. FaceForensics++ contain 1,000 real videos and 1,000 Fake videos for each type of manipulation including Face2Fcae, DeepFake and NeuralTextures. For each category of real/fake videos, the dataset was split into 720 videos for training, 140 for validation, and 140 for testing. We used videos with light compression (quantization = 23) and high compression (quantization = 40) ; this is more challenging than use of raw images and allows us to compare with a larger set of methods. Images were extracted from videos using the settings in \cite{Cozzolino2017}: 200 frames of each training video were used for training, and 10 frames of each validation and testing video were used for validation and testing respectively.

\subsubsection{Face2Face} Face2Face~\cite{Face2face} is a facial reenactment system that transfers the expressions of a source video to a target video while maintaining the identity of the target person \cite{Rossler_2019_ICCV}. The modifications brought to the target image are in the form of change of movement of the head, lips, and facial expression. 

\subsubsection{NeuralTextures} The NeuralTextures dataset \cite{NT} show facial reenactment as an example for their NeuralTextures-based rendering approach. It uses the original video data to learn a neural texture of the target person, including a rendering network. This is trained with a photometric reconstruction loss in combination with an adversarial loss. In the NeuralTextures, only the facial expressions corresponding to the mouth region is modified, when the eye region stays unchanged.

\subsubsection{DeepFake} DeepFake~\cite{Deepfake} is one of the popular face manipulation methods created using a Deep Neural Network, which swaps a facial image of a person with a different person's face, followed by editing \cite{Poisson}.

\textbf{DeepFake Video Datasets.} 
Deepfakes are a recent off-the-shelf manipulation technique that allows anyone to swap two identities in a single video. In addition to Deepfakes, a variety of GAN-based face swapping methods have also been published. The problem of deepfake detection has increased considerable attention, and this research has been stimulated with many datasets. The DFDC dataset \cite{DFDC2020} is by far the largest currently and publicly-available face swap video dataset, with over 100,000 total clips sourced from 3,426 paid actors, produced with several Deepfake, GAN-based, and non-learned methods.

We summarize and analyze seven benchmark deepfake video detection datasets in Table \ref{table1}. As Table \ref{table1} shows, all the deepfake datasets are based on face identity swapping. The only dataset contains both identity and expression manipulation is Faceforensics++ \cite{Faceforensics}. In Faceforensics++, Face2Face and NeuralTextures are the only manipulation techniques change the facial expression while two other techniques (FaceSwap and DeepFake) are based on identity change. Thus, all the other datasets contain only identity manipulated faces. Only the part of Faceforensics++ containing facial expression changes allows us to analyze facial expression manipulation. To demonstrate the difference between identity swap and expression swap, we show some examples from both categories. Fig. \ref{fig4} shows 10 frames from 4 different face manipulation datasets (Face2Face, NeuralTextures, DFDC and DeepFake). As we can see, tampered videos from Face2Face and NeuralTextures are undergo expression manipulation. The shape of lips and eyebrows which contribute substantially to facial expressions is changed in most of the frames from Face2Face vidoes. To the contrary, the tampered videos from DFDC and DeepFake do not demonstrate any major expression change in comparison to original ones. Therefore, only two datasets (Face2Face and NeuralTextures from Faceforensics++) satisfy the criteria and we evaluate the performance on those datasets. We also present the results on DeepFake datasets to demonstrate that our method performs at-par with the state-of-the-art methods in cases where the expression is not manipulated, but rather the identity is changed, thus ensuring generalizability of the approach. (Further details in Sec. \ref{results})

\textbf{Facial Expression Datasets.}
We use AffectNet~\cite{Affectnet}, a new database of facial expressions in the wild. We trained our FER system on AffectNet training set and used Face2Face, NeuralTextures and DeepFake datasets as manipulation detection datasets. AffectNet contains more than 1M facial images collected from the Internet. The dataset is divided into 11 facial expression categories - neutral, happiness, sadness, surprise, fear, disgust, anger, contempt, none, uncertain, and no-Face.

\subsection{Implementation}
In face forensics, faces play an important role and contain key features for manipulation detection. Therefore, instead of using the whole image, we extract the faces as a pre-processing step and only use the face regions to train the models.
The FER system has shared and ensemble layers consisting of CNNs using 5x5 and 3x3 convolutional windows with the stride of 1. Following each convolutional layer is a batch normalization layer \cite{batch}. 

The encoder-decoder architecture consists of XceptionNet with separable
convolution as encoder, spatial pyramid pooling module and CNNs with 3x3 convolutional windows as decoder. We transfer XceptionNet to our task by replacing the final fully connected layer with two outputs. The other layers except last convolutional layer (the layer after feature concatenation) and fully connected layer are initialized with the ImageNet weights. To set up the newly inserted fully connected layer, we fix all weights up to the last convolutional layer and pre-train the network for 3 epochs. After this step, we train the network for 20 more epochs and choose the best performing model based on validation accuracy.

The framework is implemented on PyTorch. We trained the network using the ADAM optimizer \cite{adam} with a learning rate of $0.001$, a batch size of 16, $\beta$ of $0.9$ and $0.999$, and $\epsilon$ equal to $10^{\num{-8}}$.

\begin{table*}
\centering
\caption{Classification performance in terms of accuracy for state-of-art architectures on DeepFake, Face2Face and NeuralTextures datasets with two level of video quality.}
\begin{tabular}{c|c|p{.6cm}p{.6cm}p{.6cm}|p{.6cm}p{.6cm}p{.6cm}}
\hline
\multirow{2}{*}{} & \multirow{2}{*}{Method} & \multicolumn{3}{c|}{\underline{HQ (compressed 23)}} & \multicolumn{3}{c}{\underline{LQ (compressed 40)}} \\
& & DF & F2F & NT & DF & F2F & NT  \\
\hline
\hline
&&&&&&&\\
\multirow{7}{*}{Without FER} & Steg.Features+SVM \cite{svm} & 77.12 & 74.68 & 76.94 & 65.58 & 60.58 & 60.69 \\

& Cozzolino  et  al \cite{Cozzolino2017} & 81.78 & 85.32 & 80.60 & 68.26 & 62.08 & 62.42\\

& Bayar  and  Stamm \cite{bayar} & 90.18 & 94.93 & 86.04 & 80.95 & 76.83 & 72.38\\

& Rahmouni et al \cite{rahmouni} & 82.16 & 93.48 & 75.18 & 73.25 & 67.08 & 62.59\\

& MesoNet \cite{Afchar2018MesoNetAC} & 95.26 & 95.84 & 85.95 & 89.52 & 83.56 & 75.74\\

& MultiTask \cite{Nguyen2019MultitaskLF} & 93.92 & 92.77 & 88.05 & 85.77 & 82.31 & 80.67\\

& XceptionNet \cite{xception} & 98.85 & 98.23 & 94.50 & 94.28 & 91.56 & 82.11 \\
&&&&&&&\\
\hline
\hline
&&&&&&&\\
\multirow{2}{*}{With FER} & MultiTask+EnsFER & 94.10 & 95.22 & 89.15 & 86.31 & 85.89 & 81.46 \\
&EMD (ours) & 99.13 & 99.03 & 96.31 & 95.88 & 94.45 & 83.67\\
&&&&&&&\\
\hline
\hline
\end{tabular}
\label{table2}
\end{table*}

\begin{table}[t]
\centering
\caption{Classification performance in terms of accuracy for state-of-art architectures on Face2Face datasets with two level of video quality}
\begin{tabular}{c|c|p{.9cm}|p{.9cm}}
\hline

\multirow{2}{*}{} &\multirow{1}{*}{Method} & \multicolumn{1}{c|}{HQ } & \multicolumn{1}{c}{LQ} \\

\hline
\hline
&&&\\
\multirow{2}{*}{Without FER} 
&LAE \cite{LAE} & 90.93 & -\\
&DCNN \cite{Nguyen2019MultitaskLF} & 93.50 & 82.13\\
&FT-res \cite{FT} & 94.47 & - \\
&Two-stream \cite{twoS} & 96.00 & 86.83 \\
&Capsule-Forensics \cite{capsule} & 97.13 & 81.20\\
&Face X-ray \cite{xray} & 97.73 & - \\
&&&\\
\hline
\hline
&&&\\
\multirow{2}{*}{With FER} & MultiTask+EnsFER & 95.22 & 85.89\\
&EMD (ours) & 99.03 & 94.45\\

&&&\\

\hline
\hline
\end{tabular}
\label{table3}
\end{table}

\begin{table}[t]
\centering
\caption{Segmentation performance in terms of accuracy for all evaluated architectures on Face2Face and NeuralTextures datasets with two level of video quality}
\begin{tabular}{c|c|p{.6cm}p{.6cm}|p{.6cm}p{.6cm}}
\hline
\multirow{2}{*}{} &\multirow{2}{*}{Method} & \multicolumn{2}{c|}{\underline{HQ }} & \multicolumn{2}{c}{\underline{LQ }} \\

& & F2F & NT & F2F & NT \\
\hline
\hline
&&&&\\
\multirow{2}{*}{Without FER} 

&MultiTask \cite{Nguyen2019MultitaskLF} & 90.27 & 88.67& 87.76 & 84.55\\

&XceptionNet \cite{xception} & 96.13 & 91.34 & 92.45 & 89.39\\
&&&&\\
\hline
\hline
&&&&\\
\multirow{2}{*}{With FER} & MultiTask+EnsFER & 93.22 & 90.56 & 89.31 & 86.56\\
&EMD (ours) & 98.43 & 93.78 & 95.22 & 91.54\\
&&&&\\

\hline
\hline
\end{tabular}
\label{table4}
\end{table}

\begin{table}[t]

\centering
\caption{Classification (cls) and segmentation (seg) accuracy for different FER architectures on Face2Face and NeuralTextures datasets with high quality videos.}
\begin{tabular}{c|p{.6cm}p{.6cm}|p{.6cm}p{.6cm}}
\hline
\multirow{2}{*}{Method} & \multicolumn{2}{c|}{\underline{Face2Face}} & \multicolumn{2}{c}{\underline{NeutalTexture)}} \\
& Cls & Seg & Cls & Seg \\
\hline
\hline
&&&&\\
MultiTask+ SimFER & 94.93 & 91.84& 88.62 & 89.21\\

XceptionNet+ SimFER & 98.63 & 96.89 & 95.83 & 92.44\\

MultiTask+EnsFER & 95.22 & 93.22 & 89.15 & 90.56 \\

EMD (ours) & 99.03 & 98.43 & 96.31 & 93.78 \\
&&&&\\

\hline
\hline
\end{tabular}
\label{table5}
\end{table}

\subsection{State-of-the-art Methods}
\subsubsection{Manipulation detection.}
Manipulation detection is a classification problem where the methods focus on identifying whether or not an image or video is manipulated. Some of the prominent works in this field that we compare against are as follows: 
\begin{itemize}
\item\textbf{Steg. Features+SVM} \cite{svm} is based on steganalysis and employs handcrafted features. The features are co-occurrences on 4 pixels patterns along the horizontal and vertical direction on the high-pass images for a total feature length of 162. These features are then used to train a linear Support Vector Machine (SVM) classifier.

\item\textbf{Cozzolino et al.} \cite{Cozzolino2017} combined the hand-crafted steganalysis features from \cite{svm} with a CNN-based network. 

\item\textbf{Bayar and Stamm} \cite{bayar} proposed a CNN that uses a constrained convolutional layer followed by two convolutional, two max-pooling and three fully-connected layers. The constrained convolutional layer is specifically designed to suppress an image’s content and learn manipulation detection features.

\item\textbf{Rahmouni et al.} \cite{rahmouni} proposed a CNN architecture with a custom pooling layer to optimize current
best-performing algorithms' feature extraction scheme. 

\item\textbf{MesoNet} \cite{Afchar2018MesoNetAC} is a CNN-based network using InceptionNet [56] in face forensics area. The network has two inception modules and two convolution layers with max-pooling, followed by two fully-connected layers. Instead of the cross-entropy loss, the authors use the mean squared error between true and predicted labels.

\item\textbf{XceptionNet} \cite{xception} uses a deep neural network trained on ImageNet. This architecture is constructed by modifying the inception modules where the depthwise separable convolution is used. There are a total of 36 convolutional layers used in the base network. \cite{Rossler_2019_ICCV} transfered it to the manipulation detection task by replacing the final fully connected layer with two outputs. The other layers are initialized with the ImageNet weights.

\item\textbf{LAE} \cite{LAE}, \textbf{FT-res} \cite{FT} both attempting to learn intrinsic representation instead of capturing artifacts in the training set.

\item\textbf{DCNN} \cite{dcnn} proposes an approach leveraging the transferable features from a pre-trained Deep Convolutional Neural Networks (D-CNN) to detect manipulation in face images.

\item\textbf{Two-stream} \cite{twoS} uses a two-stream CNN to achieve higher performance in image forgery detection. They use standard CNN network architectures to train the model.

\item\textbf{Capsule-Forensics} \cite{capsule} presents a method that uses a capsule network to detect forged images and videos.


\item\textbf{Face X-ray} \cite{xray} proposes a face forgery detection method based on the observation that most existing face manipulation methods share a common blending step and there exist intrinsic image discrepancies across the blending boundary.

\end{itemize}

\subsubsection{Manipulation Segmentation}
In order to spatially localize the exact position of manipulation in an image, which in this case is part of a face, we use a decoder in our proposed architecture to segment out the tampered region. FaceForensics \cite{Faceforensics} and FaceForensics++ \cite{Rossler_2019_ICCV} datasets  provide binary mask to show the manipulation region in the facial images. 
The most closely related work which also performs segmentation with classification is the following.
\begin{itemize}
\item\textbf{MultiTask} \cite{Nguyen2019MultitaskLF}
outputs both the probability of an image being spoofed and segmentation maps of the manipulated regions in each frame of the input. For this method a CNN is designed that uses the multi-task learning approach to simultaneously detect manipulated images and videos and locate the manipulated regions for each query. Information gained by performing one task is shared with the other task.
\end{itemize}

\subsection{Quantitative Comparisons}
\textbf{Evaluation Metrics.} 
In terms of evaluation metrics, we use classification accuracy for the manipulation detection, which represents how many test images are correctly classified. For segmentation tasks, we use pixel-wise classification accuracy which indicates whether a pixel in an image is manipulated or not.

\subsubsection{Results.}
\label{results}
Table \ref{table2} shows the classification accuracy of different methods for three Deepfake (DF), Face2Face (F2F) and NeuralTextures (NT) datasets using two types of video quality (low quality (LQ) and high quality (HQ)).  As may be observed, in terms of  classification accuracy, EMD (our method) reaches the best performance on all three datasets. In comparison to XceptionNet, our proposed method achieves $\sim{3\%}$ and $\sim{2\%}$ improvements in classification accuracy on F2F and NT datasets respectively with low quality videos. 

We add FER system to MultiTask \cite{Nguyen2019MultitaskLF} architecture which also leads to improvements of accuracy by $\sim{3\%}$ and $\sim{1\%}$ on F2F and NT datasets with high quality videos. More improvement in classification accuracy can be seen on F2F and NT dataset where the facial expression is manipulated. However, we observe that addition of FER system has small affect on Deepfake detection which does not have expression manipulation, but there is no fall in performance. This demonstrates the generalizability of our approach. Furthermore, we compare our method with more of the state-of-art methods on Face2Face dataset in terms of classification accuracy. Table \ref{table3} shows this comparison. As it is clear from Table \ref{table3}, our method achieves higher classification accuracy.

For localization task, Table \ref{table4} shows $\sim{3\%}$ improvement of segmentation accuracy on low quality videos from F2F dataset and $\sim{2\%}$ imporvement on NT dataset with the same video quality. 

\subsubsection{Ablation study} To demonstrate the effectiveness of utilizing FER in manipulation detection and segmentation, we run different experiments with variation of FER architecture. As we can see from Table \ref{table5}, using FER system with multiple branches and selecting the most informative feature maps by using ESR (the one we use in our architecture), achieves higher accuracy in both detection and segmentation task . Using simple FER (SimFER) consisting of shallow convolutional layers leads to performance drop by $\sim{1\%}$ and $\sim{2\%}$ in classification and segmentation for both F2F and NT datasets with high quality videos.

\subsection{Analysis of Results}
\subsubsection{Effect of FER on manipulation detection.}
\begin{figure}[ht]
\includegraphics[width=\linewidth]{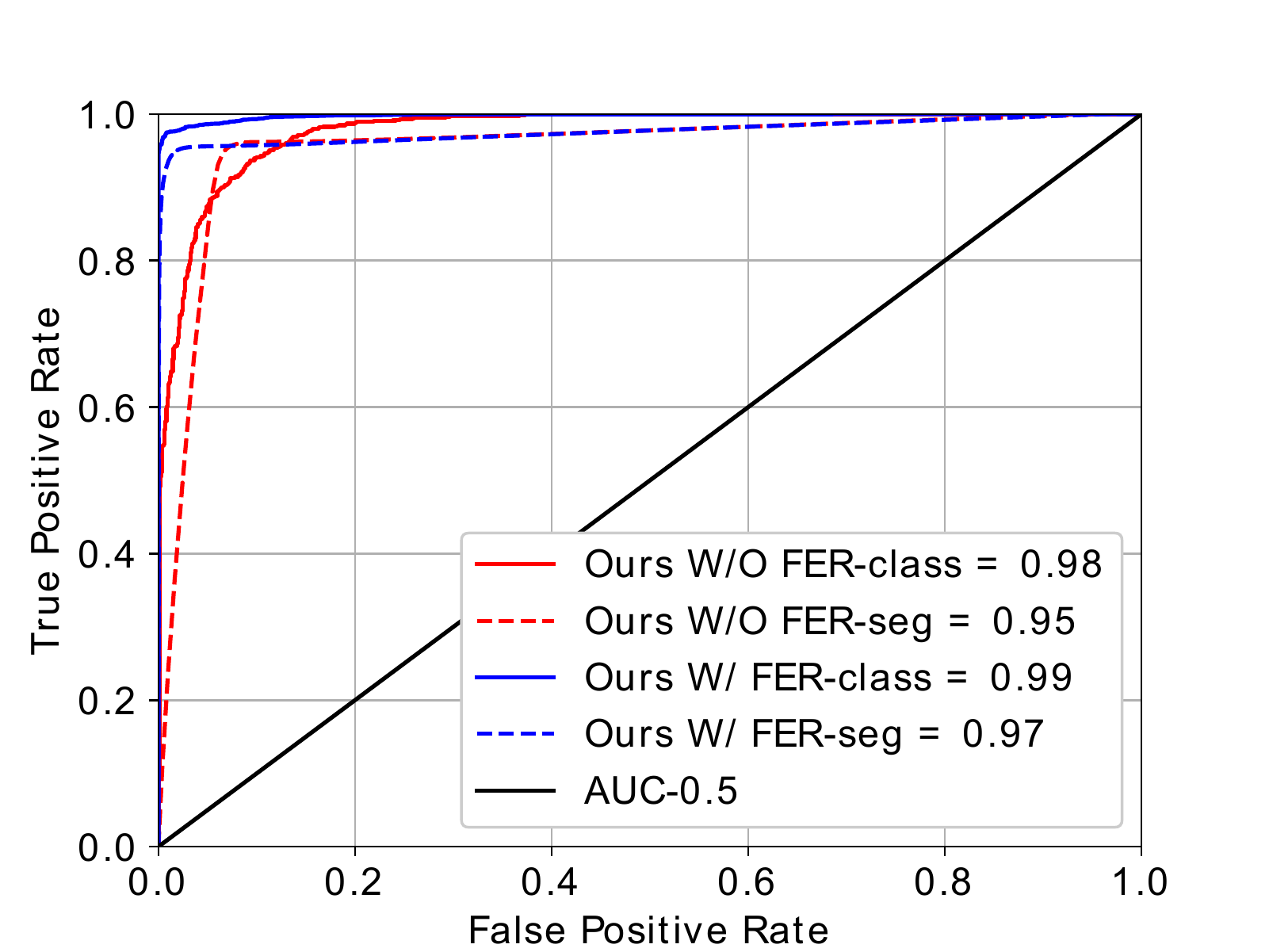}
\caption{ROC curves for classification and segmentation with and without facial expression recognition stream on F2F dataset. The solid and dotted blue lines are the proposed EMD algorithm.} \label{fig5}
\end{figure}

\begin{figure}[ht]
\includegraphics[width=\linewidth]{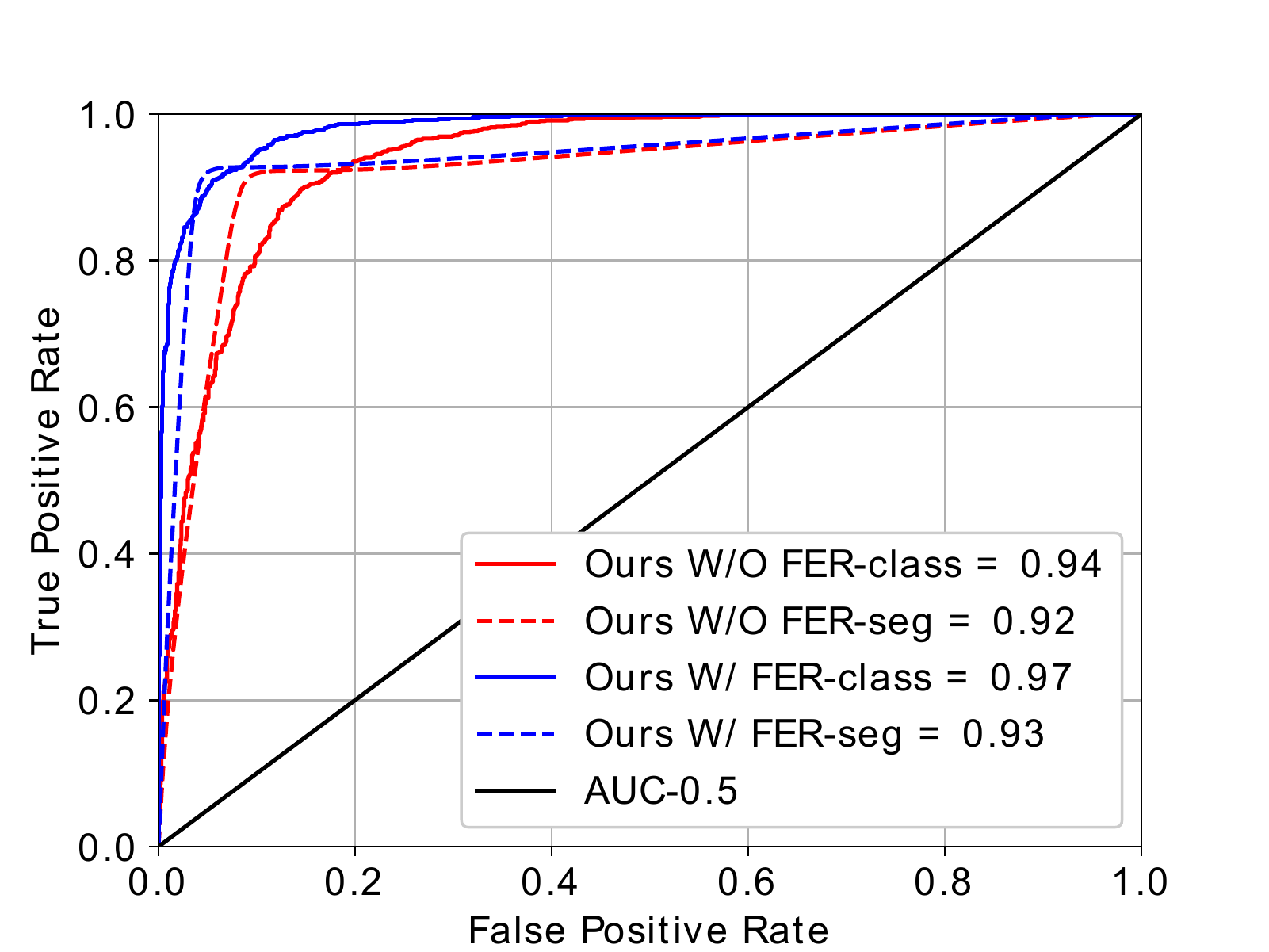}
\caption {ROC curves for classification and segmentation with and without facial expression recognition stream on NT dataset.The solid and dotted blue lines are the proposed EMD algorithm.} \label{fig6}
\end{figure}

We use ROC curves to show the benefit of FER in manipulation detection. Fig. \ref{fig5} and Fig. \ref{fig6} demonstrate ROCs for both detection and segmentation tasks with and without FER system. AUC score for our network with FER stream (EMD) achieves 99\% and 97\% for detection and segmentation tasks on F2F respectively. Thus, our method leads to $\sim{1\%}$ and $\sim{2\%}$ improvement in detection and segmentation AUC score in comparison to its counterpart without the FER stream. Based on Fig. \ref{fig6}, our method achieves $\sim{3\%}$ and $\sim{1\%}$ improvement in detection and segmentation AUC score when it is trained/tested on NT dataset.

\begin{figure}[ht]
\includegraphics[width=\linewidth]{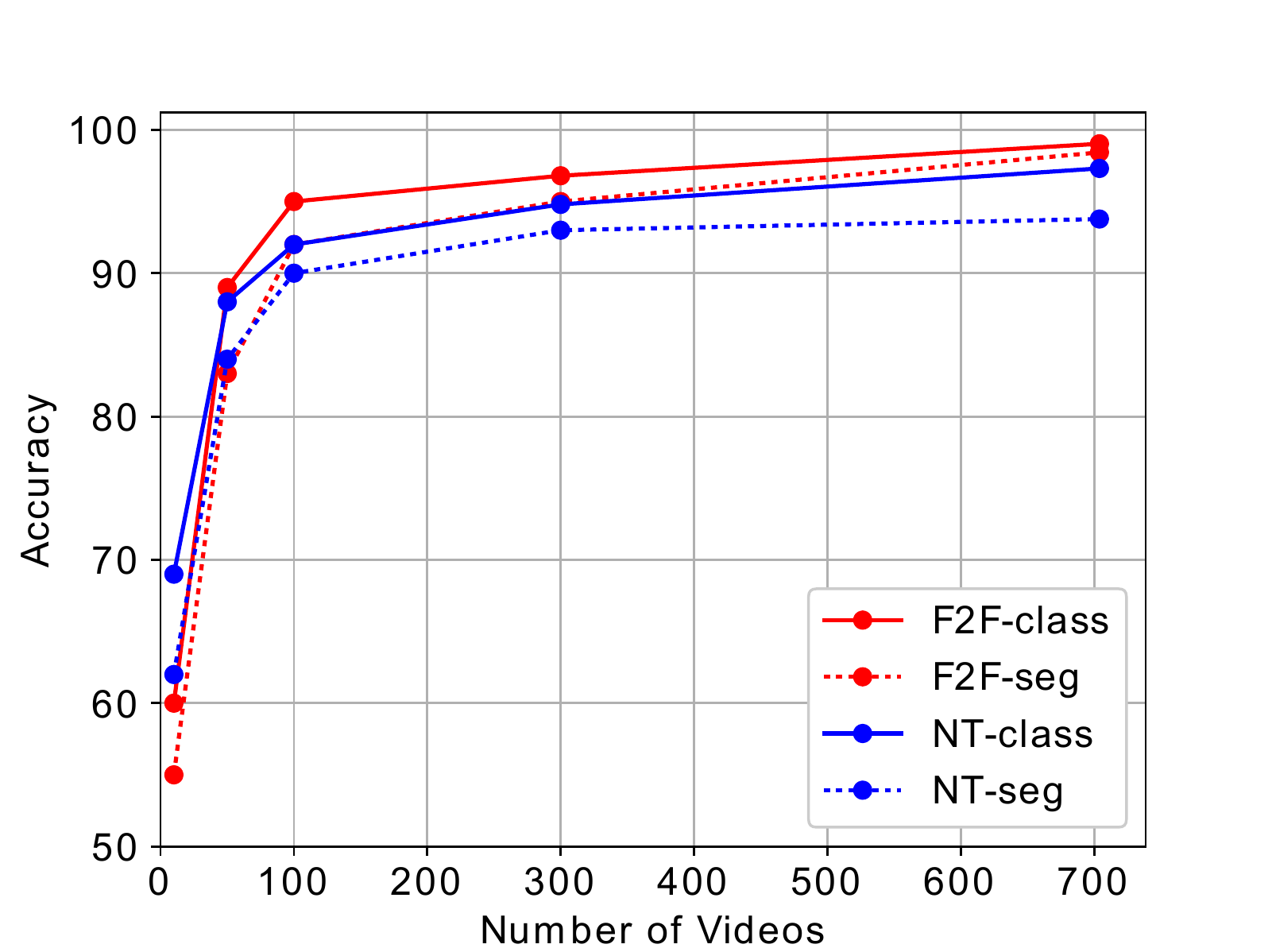}
\caption {The detection and segmentation performance of our approach by varying the training corpus size on F2F and DF datasets.} \label{fig7}
\end{figure}

\subsubsection{Size of training data.} As shown in Fig. \ref{fig7}, we evaluate our proposed network on training sets with variable sizes. This evaluation demonstrates that our method even performs well when there is not much data available to train. For both datasets (F2F and NT), we compute classification and segmentation accuracy varying the training size from $10$ to $\sim{700}$ videos. As we can see from the figures, by adding more videos to the datasets our performance increases initially. Adding more than 300, our model's performance does not change much indicating our method can perform good enough with even $\sim{300}$ videos.

\begin{figure*}[t]
\includegraphics[width=\linewidth]{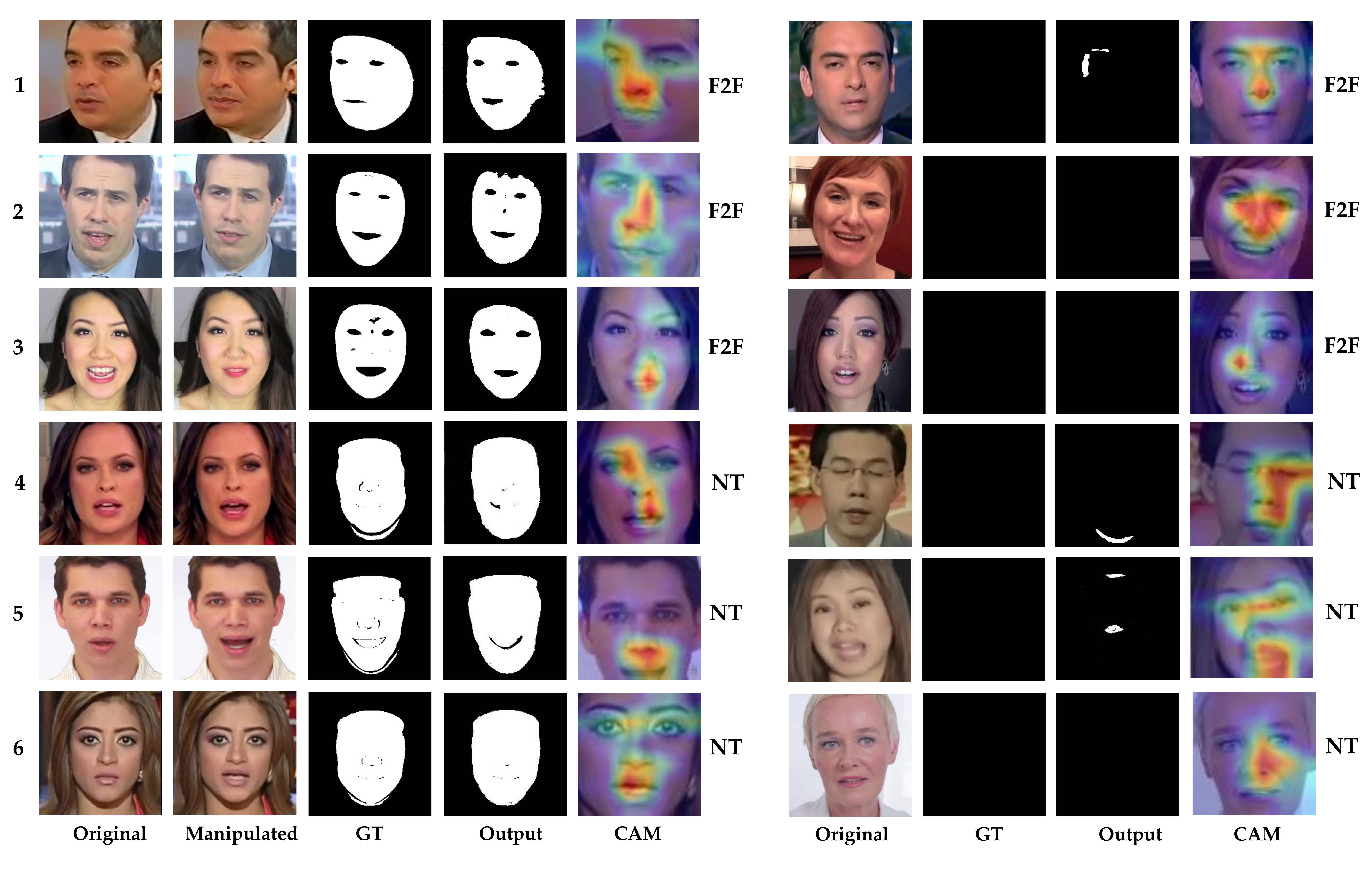}
\caption{First and second columns of the left section show the original images and manipulated ones respectively. The black and white images in the third column are corresponding binary GT masks. Predicted masks (column 4) and generated class activation maps (column 5) for manipulated images from Face2Face (row 1,2,3) and  NeuralTextures (row 4,5,6) dataset. In the right section of the figure, we show pristine images with their corresponding binary GT masks, predicted masks and CAMs.} \label{fig3}
\end{figure*}

\subsubsection{Why does FER help?} 
To show the effect of the features extracted from FER system, we visualize the last layer of CNN in our FER. For this purpose, we compute CAMs. A CAM for a particular category indicates the discriminative image regions used by the CNN to identify that category. Work by \cite{Zhou2014ObjectDE} has shown that the convolutional  units  of  various  layers  of CNNs  actually  behave  as  object  detectors  despite no supervision on the location of the object provided. 

In fact, the network can retain its remarkable localization ability until the final layer. This feature allows identification of the discriminative image regions which are important for manipulation detection. Specifically, for expression change detection, addition of a network which can localize regions in the face with information about the expressions helps manipulation detection methods to perform better. 

As Fig. \ref{fig3} represents, expression changes happen mostly around eyes, mouth and eyebrows. In the last column of both right and left section of figure, we generate the CAMs for manipulated and pristine images in F2F and NT datasets. As it is clear, our network can classify expressions quite well although the main FER stream has been trained on a different dataset (AffectNet).

\section{Conclusions}
In this paper, we propose a new approach (EMD) to exploit facial expression systems in image/video facial expression manipulation detection. Application of deep network layers rich in information about facial expressions improves the manipulation detector by making it learn the useful features for facial expression transformation. Experiments on two challenging datasets demonstrate our method has better classification and segmentation performance in facial expression manipulation detection in comparison to state-of-art results. Also, our method is close to the state-of-the-art methods for other kinds of manipulation detection, thus ensuring generalizability. 


\ifCLASSOPTIONcaptionsoff
  \newpage
\fi

\bibliographystyle{IEEEtran}
\bibliography{bare_jrnl}

\begin{IEEEbiography}[{\includegraphics[width=1.1in,height=1.1in,clip]{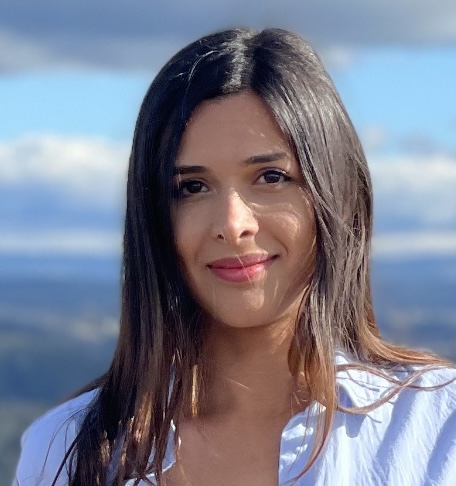}}]{Ghazal Mazaheri} received her Bachelor’s degree in Electrical and Electronic Engineering from
K.N.Toosi University of Technology, Tehran in 2017. She is currently pursuing her Ph.D. degree in the department of Computer Science and Engineering at University of California, Riverside. Her main research interests include computer vision, machine learning and multimedia forensics.

\end{IEEEbiography}

\begin{IEEEbiography}[{\includegraphics[width=1.1in,height=1.1in,clip]{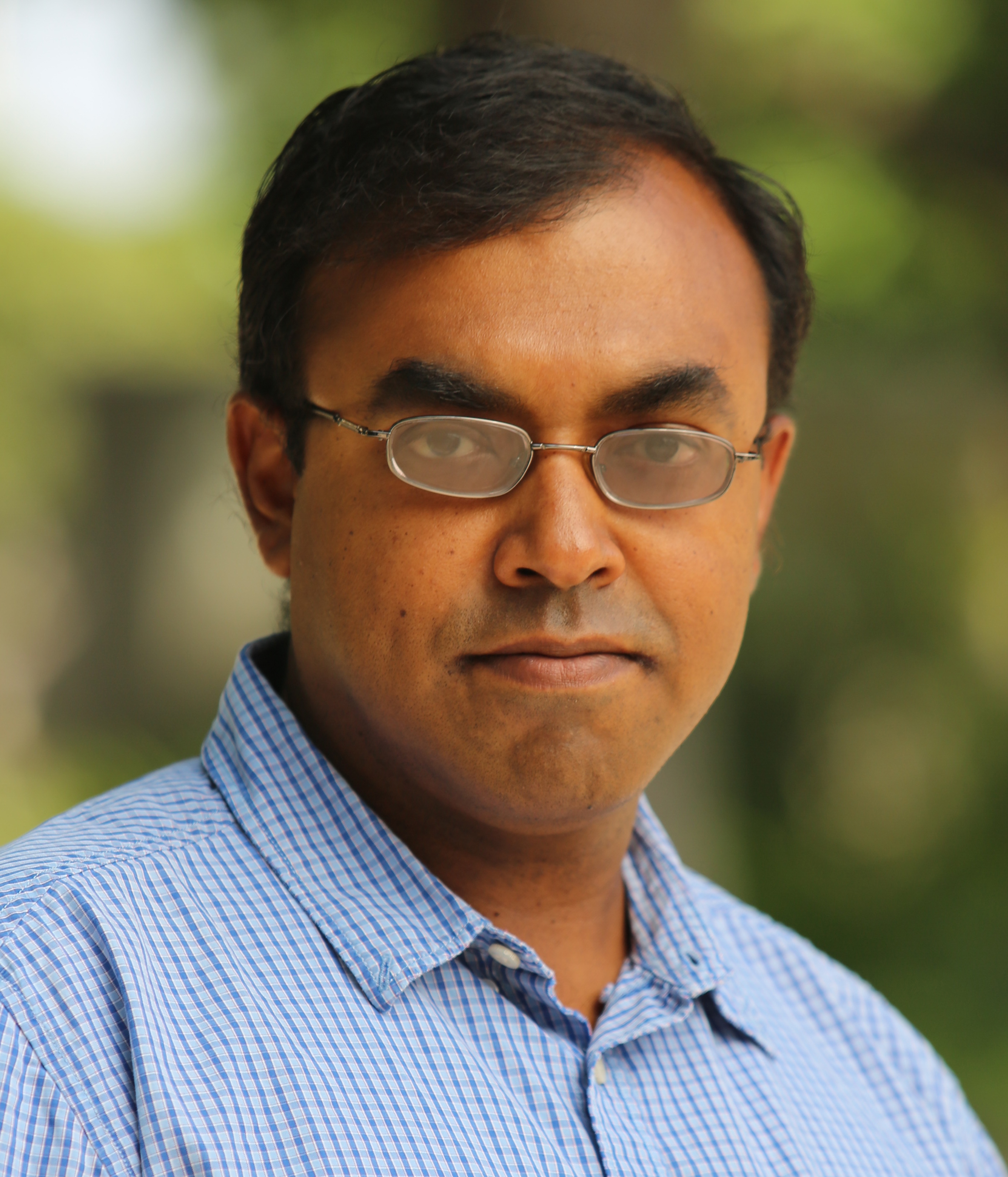}}]{Amit Roy-Chowdhury} received his PhD from the University of Maryland, College Park (UMCP) in Electrical and Computer Engineering in 2002 and joined the University of California, Riverside (UCR) in 2004 where he is a Professor and Bourns Family Faculty Fellow of Electrical and Computer Engineering, Director of the Center for Robotics and Intelligent Systems, and Cooperating Faculty in the department of Computer Science and Engineering. He leads the Video Computing Group at UCR, working on foundational principles of computer vision, image processing, and vision-based statistical learning, with applications in cyber-physical, autonomous and intelligent systems. He has published about 200 papers in peer-reviewed journals and conferences. He is the first author of the book Camera Networks: The Acquisition and Analysis of Videos Over Wide Areas. His work on face recognition in art was featured widely in the news media, including a PBS/National Geographic documentary and in The Economist. He is on the editorial boards of major journals and program committees of the main conferences in his area. His students have been first authors on multiple papers that received Best Paper Awards at major international conferences, including ICASSP and ICMR. He is a Fellow of the IEEE and IAPR, received the Doctoral Dissertation Advising/Mentoring Award 2019 from UCR, and the ECE Distinguished Alumni Award from UMCP.
\end{IEEEbiography}

\end{document}